# Developing an efficient corpus using Ensemble Data cleaning approach

Dr Md Taimur Ahad (Corresponding Author)
Associate Professor
Computer Science and Engineering
Daffodil International University
Savar, Bangladesh
email: taimurahad.cse@diu.edu.bd
mdtaimurahad@gmail.com

**Abstract**

*Despite the observable benefit of Natural Language Processing (NLP) in processing a large amount of textual medical data within a limited time for information retrieval, a handful of research efforts have been devoted to uncovering novel data-cleaning methods. Data cleaning in NLP is at the centre point for extracting validated information. Another observed limitation in the NLP domain is having limited medical corpora that provide answers to a given medical question. Realising the limitations and challenges from two perspectives, this research aims to clean a medical dataset using ensemble techniques and to develop a corpus. The corpora expect that it will answer the question based on the semantic relationship of corpus sequences. However, the data cleaning method in this research suggests that the ensemble technique provides the highest accuracy (94%) compared to the single process, which includes vectorisation, exploratory data analysis, and feeding the vectorised data. The second aim of having an adequate corpus was realised by extracting answers from the dataset. This research is significant in machine learning, specifically data cleaning and the medical sector, but it also underscores the importance of NLP in the medical field, where accurate and timely information extraction can be a matter of life and death. It establishes text data processing using NLP as a powerful tool for extracting valuable information like image data.*

## 1. Introduction

Natural Language Processing (NLP) has attracted considerable attention as a research topic in the medical domain as NLP effectively produces valuable information from patients' records and doctors' transcriptions. NLP is extended to the medical field as many analysts advocated that NLP has proven to be an effective data mining tool to extract the required information from unstructured medical data Nancy & Maheswari (2020), Ahmed et al. (2020). Medical texts and records contain a large amount of applicable information, and NLP as an intelligent data mining tool is capable of extracting valuable knowledge such as finding better medicines, predicting epidemics and finding new genome sequences Deshmukh & Phalnikar (2021), Serrano et al. (2020), Jusoh (2018).

However, data cleaning is necessary for unstructured medical data, as unstructured data are challenging to process and generate meaningful information. Moreover, unstructured data are complex to process using machines Cambria & White (2014). In general, unstructured data cleaning can be described as producing clean data Jermyn et al. (1999). Usually, unstructured data cleaning requires the removal of characters (stop words, punctuation, special characters, links, @ as these data are not necessary for producing information Gull et al. (2017).

One NLP data cleaning approach, namely the ensemble technique, has been proven efficient in cleaning unstructured data (Sangamnerkar et al., 2020; Fattahi & Mejri; Araque & Iglesias, 2021). An ensemble is a meta-algorithm that combines several base models into one predictive model, and this combination has shown supreme performance in many machine-learning tasks (Liu et al., 2019). The ensemble technique uses text summarisation algorithms as a filter to produce critical content from the targeted text file or records in an Excel file. The fundamental working method is that text or records are taken as input for processing, the ensemble technique algorithm is implemented, and information is generated as meaningful words or sentences (Duan et al., 2021).

Recently, a new approach in NLP is an attempt to develop a medical corpora that retrieves medical information, which has been a research topic. A corpus (corpora) is a large and structured set of machine-readable texts produced in an NLP setting Karami et al. (2018). The NLP researchers suggest that Knowledge Engineering largely depends on using corpora as an answer bank to the knowledge access problem in the medical domain. The researchers propose to fulfil a particular objective (a medical-related question), and the corpus guarantees the reliability and stability of the answer/s Baneyx et al. (2007). However, developing an efficient corpus for discovering the hidden structure in significant unstructured health and medical data sets is challenging (Karami et al., 2018). Karami et al. (2018) complained that substantial resources were allocated to developing new data analytic methods and tools; retrieving extensive health and medical data from a corpus is still challenging (Tian et al., 2019).

Patient datasets are increasing daily as hospitals and medical centres are digitalising continuously; anomalies in patient data, such as incompleteness, inconsistency, and value contradicting, are observable in the patient records. This affects data analysis and thus possibly impacts valuable information Iyer et al. (2015). According to a study, data scientists spend 60% of their time cleaning data when processing enormous amounts of data (CrowdFlower,  Data scientists also predict that data cleaning consumes 40 to 50% of an organisation's budget due to its time-consuming nature (2013). An intelligent end-to-end data cleaning can save organisations and researchers much money. Moreover, an efficient corpus can provide users with valuable information.

Realising the limitations and challenges from two perspectives, firstly, challenges in cleaning medical datasets and secondly, limited efficient medical corpora, this research aims to clean a medical dataset using ensemble techniques and to develop a corpus that answers a medical-related question from the unstructured dataset. This research expects the corpora to answer the question based on the semantic relationship of corpus sequences. To fulfil the aim, a precise methodological process in NLP was undertaken to build various types of medical ontologies according to the differential semantics principles of a medical term.

# 2. Literature review

The literature review suggests two main utilisations of Ensemble techniques are observed; firstly, a popular method is utilising ensemble techniques to increase the accuracy in detecting or predicting diseases, and secondly, few papers utilise ensemble as a data cleaning tool. We first start with the first technique:

Among the first group that utilised ensemble as a tool to improve accuracy and prediction, there is a joint agreement among researchers that a heterogeneous ensemble or multi-label method via ensemble techniques yields considerable performance compared to single-label methods et al. (2012). Following utilising a multi-label ensemble to increase the dependent variable's accuracy, Bashir et al. (2014) developed a framework to detect patients' heart disease using a vote-based classifier. In the study by Bashir et al. (2014), ensemble scheme performances were compared with other individual classifiers such as Neural Network, Decision Tree, Na¨ıve Bayes, Support Vector Machine, Perceptron and AutoMLP. The average accuracy of the proposed ensemble vote scheme observed was much better than that of other techniques. It has achieved an average accuracy of 83% bashir2014ensemble. Ali et al. presented an intelligent healthcare monitoring framework using an ensemble deep learning model and feature fusion methods to improve the accuracy of heart disease prediction. The prediction assists physicians in quickly and accurately diagnosing heart patients. The researchers' ensemble deep learning model effectively handles two different data sources and enhances the performance of heart disease diagnoses.

Another ensemble prediction approach by Joshi et al. concentrated on predicting diabetics prediction Joshi & Alehegn (2017). On 768 data sets from the Pima Indian Diabetes Data Set, Joshi et al. applied traditional predictive algorithms such as KNN, Na¨ıve Bayes, Random Forest, and J48. However, Joshi et al. created a hybrid model by integrating various algorithms into one to improve performance and accuracy Joshi & Alehegn (2017). An exciting study by Le et al. (2020) employed an ensemble deep neural network to create a supervised learning model. With sensitivity, specificity, accuracy, Matthews correlation coefficient (MCC), and area under the receiver operating characteristic curve (AUC) values of 60.2 per cent, 84.6 per cent, 76.3 per cent, 0.449, and 0.814, the approach could identify essential genes. Like Joshi & Alehegn (2017) and Tahir et al. (2012), the research outperformed single models without an ensemble. The results showed that the proposed method effectively determines essential genes in particular and other sequencing issues in general Le et al. (2020).

The study by Liu et al. (2019) also advocated that combining multiple NLP systems demonstrates that ensemble methods can improve the performance for both concept recognition and patient-specific concept recognition tasks and, more importantly, result in superior performance. The study claimed that manually curating standardised phenotypic concepts like Human Phenotype Ontology (HPO) terms from narrative text in electronic health records (EHRs) takes time and effort.

However, NLP assists automated phenotype extraction, making curating clinical phenotypes from clinical texts more efficient. While individual NLP systems can be effective for a single cohort, an ensemble-based approach could improve the recognition of HPO-based phenotype concepts in clinical notes Liu et al. (2019).

In the case of skin cancer detection and prediction, the ensemble method used on Dermatology datasets gives better performance as compared to different classifier algorithms Verma et al. (2019)

.The ensemble method gives more accurate and effective skin disease prediction. The researchers used five different data mining techniques and then developed an ensemble approach comprising all five data mining techniques as a single unit. The results show that the dermatological prediction accuracy of the test data set is increased compared to a single classifier.

Now, we focus on the studies that utilised ensemble as a data cleaning tool: The study by Chen et al. (2018) used the ensemble empirical mode decomposition (EEMD) method for anomaly detection and cleaning of highway elevation data extracted from Google Earth (GE). Three interstate highway segments were studied, and typical raw GE elevation data anomalies were identified. The EEMD method was then applied to decompose elevation data into different compositions with different details of original data, which were determined into those containing accurate information or white noise. The modelling results showed that the EEMD method effectively excluded noises and obtained precise elevation data. The findings of this study can help transport authorities create an accurate elevation data set for all highways or other road classes.

Perhaps an exciting use of ensemble learning algorithms used in AI-based technology by Mahdavi et al. (2019). The authors argued that iterative or manual data cleaning for unknown datasets is inappropriate as knowing the data quality problems upfront is unrealistic. Realising the issues of the unknown dataset, the study presented an automated data cleaning system using the ensemble learning method stacking. Stacking is an approach for training a meta-learner by combining multiple models for error detection.

Another ensemble learning application in power-related research was by Zhang et al. (2020). The abnormal and missing data in the original power condition monitoring dataset was cleaned based on grey correlation analysis and ensemble learning. After the grey correlation analysis method is applied to select the parameters with a high correlation degree, a data-cleaning model based on the ensemble learning method (random forest) is established. The example shows that the method presented in this paper can correctly identify many abnormal data and fill in the missing data accurately. Data quality after cleaning is improved, which benefits data mining, condition assessment and fault diagnosis for power equipment.

Now, we focus on the research that aimed to build an adequate corpus that acts as an answer bank for providing a medical-related question.

The author of the study, Abacha Zweigenbaum (2015), developed a web-based medical question-answering system combining NLP techniques. Question Answering (QA) aims to provide precise and quick answers to user questions from a collection of documents or a database. The authors addressed problems in the medical domain where several specific conditions are met. The QA was based on a semantic approach, meaning "Answer Search" was based on semantic search and query relaxation. Oracle describes query relaxation as an application that uses a query template to execute the most restrictive version of a query first, progressively relaxing the query until the required number of hits is obtained. The system performance on fundamental questions and answers shows promising results and suggests that a query-relaxation strategy can further improve the overall performance.

Studies also focused on native language-based medical corpora. For example, the survey by Cherednichenko et al. (2020) developed a Medical Corpus in Ukrainian. The Ukrainian-language corpus (UKRMED), which contains a variety of medical text genres (Clinical protocols, Blogs, and Wikipedia), describes the process of collecting, creating and processing a corpus of medical data in Ukrainian. The authors provided statistical characteristics of the corpus, and an analysis of the morphological parts of speech was provided. Frequency lemmas for this medical corps are analysed. The UKRMED corpus can be used to solve the task of natural language simplification. Another study in the Chinese context, the study of He et al. (2017), built a comprehensive corpus covering syntactic and semantic annotations of Chinese clinical texts with corresponding annotation guidelines and methods as well as developed tools trained on the annotated corpus, which supplies baselines for research on Chinese texts in the clinical domain.

# 3. Experimental description

The research adopted a qualitative methodology as the methodology is suitable when there is a need for explorative research to understand complex computing phenomena (Strauss & Corbin, 1998). The experimental design of the research is organised as follows: First, a testbed dataset of patient records is obtained from publicly available respiratory data. Second, a new data-cleaning method, the ensemble technique, is applied to the dataset. Third, the results of the experiments are reported by presenting some performance measures in graphical or table form. Following that, conclusions are drawn on the advantages of the new methodology (here, ensemble), stating that the newly developed method is the research’s essential contribution to improved performance (Eiben & Jelasity, 2002). The experiment process is depicted in Figure 1.

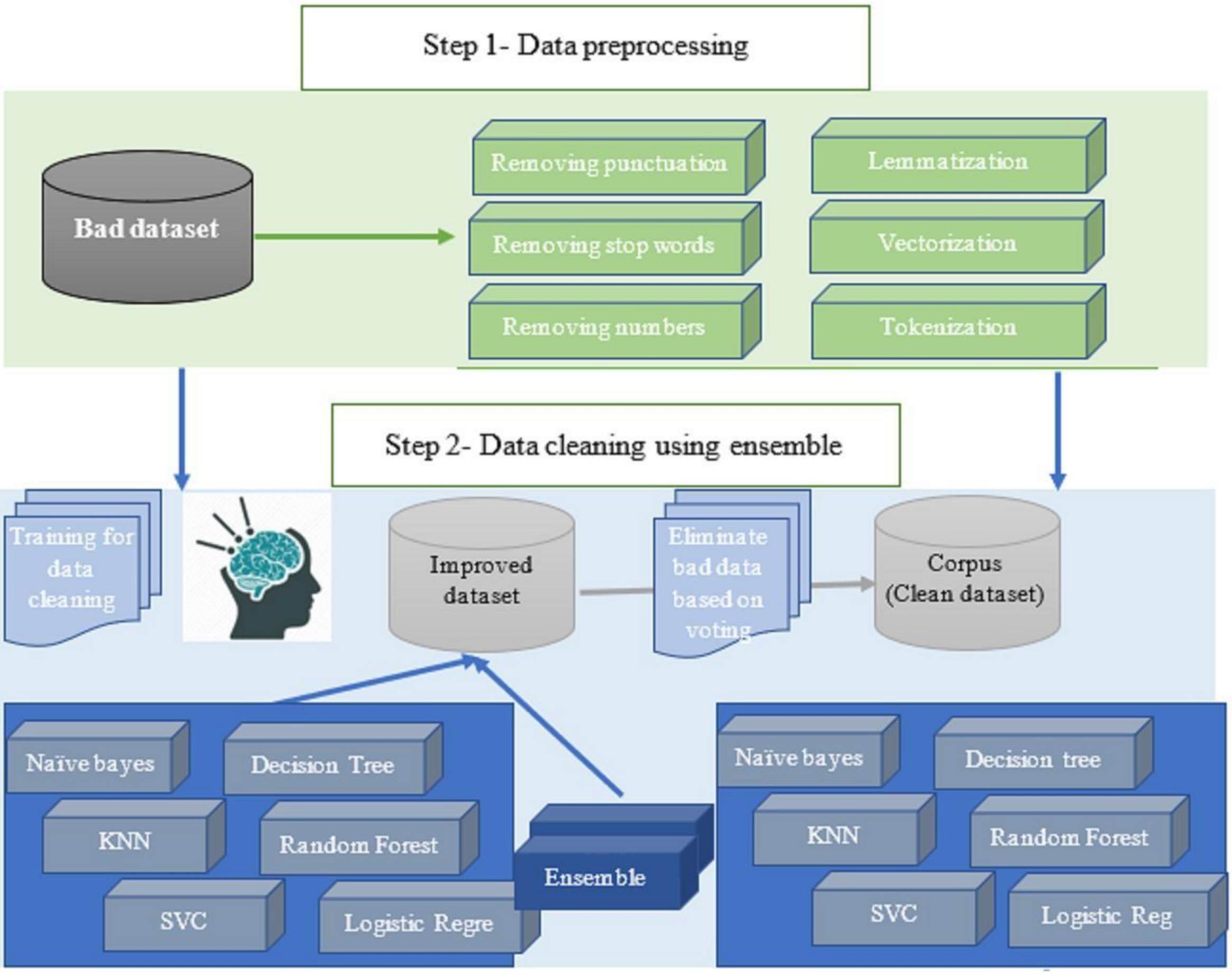

Figure 1: The initial visualisation of the unstructured/downloaded dataset

The experiment was conducted on a Jupiter notebook using Python. In this study, two datasets were considered for the experiment. The first dataset was a medical transcription, and the format was comma-separated values (CSV). The visualisation of the dataset suggested that the dataset had 4999 data points of 40 classes. We used the medical_speciality column as a predictable (dependent) column and the transcription column as an input (independent) column. The second experiment included a dataset from health care services, from basic medical diagnostics to critical emergency services, and the dataset contains 53933 data points of 8 classes. In this case, we have used the category column as a predictable (dependent) and the summary column as an input (independent).

| description | medical_specialty | sample_name | transcription | keywords |
|---|---|---|---|---|
| A 23-year-old white female presents with comp... | Allergy / Immunology | Allergic Rhinitis | SUBJECTIVE:, This 23-year-old white female pr... | allergy / immunology, allergic rhinitis, aller... |
| Consult for laparoscopic gastric bypass. | Bariatrics | Laparoscopic Gastric Bypass Consult - 2 | PAST MEDICAL HISTORY:, He has difficulty climb... | bariatrics, laparoscopic gastric bypass, weigh... |
| Consult for laparoscopic gastric bypass. | Bariatrics | Laparoscopic Gastric Bypass Consult - 1 | HISTORY OF PRESENT ILLNESS: , I have seen ABC ... | bariatrics, laparoscopic gastric bypass, heart... |
| 2-D M-Mode. Doppler. | Cardiovascular / Pulmonary | 2-D Echocardiogram - 1 | 2-D M-MODE: , ,1. Left atrial enlargement wit... | cardiovascular / pulmonary, 2-d m-mode, dopple... |
| 2-D Echocardiogram | Cardiovascular / Pulmonary | 2-D Echocardiogram - 2 | 1. The left ventricular cavity size and wall ... | cardiovascular / pulmonary, 2-d, doppler, echo... |

Figure 2: The initial visualisation of the unstructured/downloaded dataset

After removing records with null data from both datasets, the data visualisation suggested that the data distribution in both datasets was imbalanced and insufficient for each class. Balancing was, therefore, required. The data distribution (see Figure 2) for the first dataset is Surgery (1088), Consultation & Physiology (516), Cardiovascular & Pulmonary (371), and Orthopedic (355), etc, and for the second dataset is prescription (14500), appointments (12960), ask__doctor (11743), and miscellaneous (10460), etc. In the experiment, we considered the classes with at least three hundred fifty-five (355) records for the first and 10460 for the second datasets. Following the benchmark, other classes were discarded due to insufficient data for analysis.  Finally, each dataset's classes were balanced to the minimum data point, 355 data points for each class for the first dataset and 10460 data points for each class for the second dataset, using pandas. Therefore, we now have datasets of four (4) classes and two (2) columns. However, the output class was converted to numerical values between 0, 1, 2, and 3 for each class and added as a column to the dataset for learning purposes. Thus, we found the initial dataset for both experiments, and our goal is to find a new dataset that performs better on machine learning models.

The experiment on both datasets included in the research is divided into four broad sections: Firstly, a data-cleaning approach was adopted. Secondly, baseline models such as naïve Bayes, KNN, SVM, decision tree, random forest, and logistic regression were created and recorded the baseline performance for each model on the initial dataset; thirdly, an ensemble technique, using already built models such as naïve Bayes, KNN,

SVM, decision tree, random forest, logistic regression, and support vector machine, was applied to clean the original dataset, thus creates the new dataset, finally, find the performance on the new dataset. The experimental processes are described below:

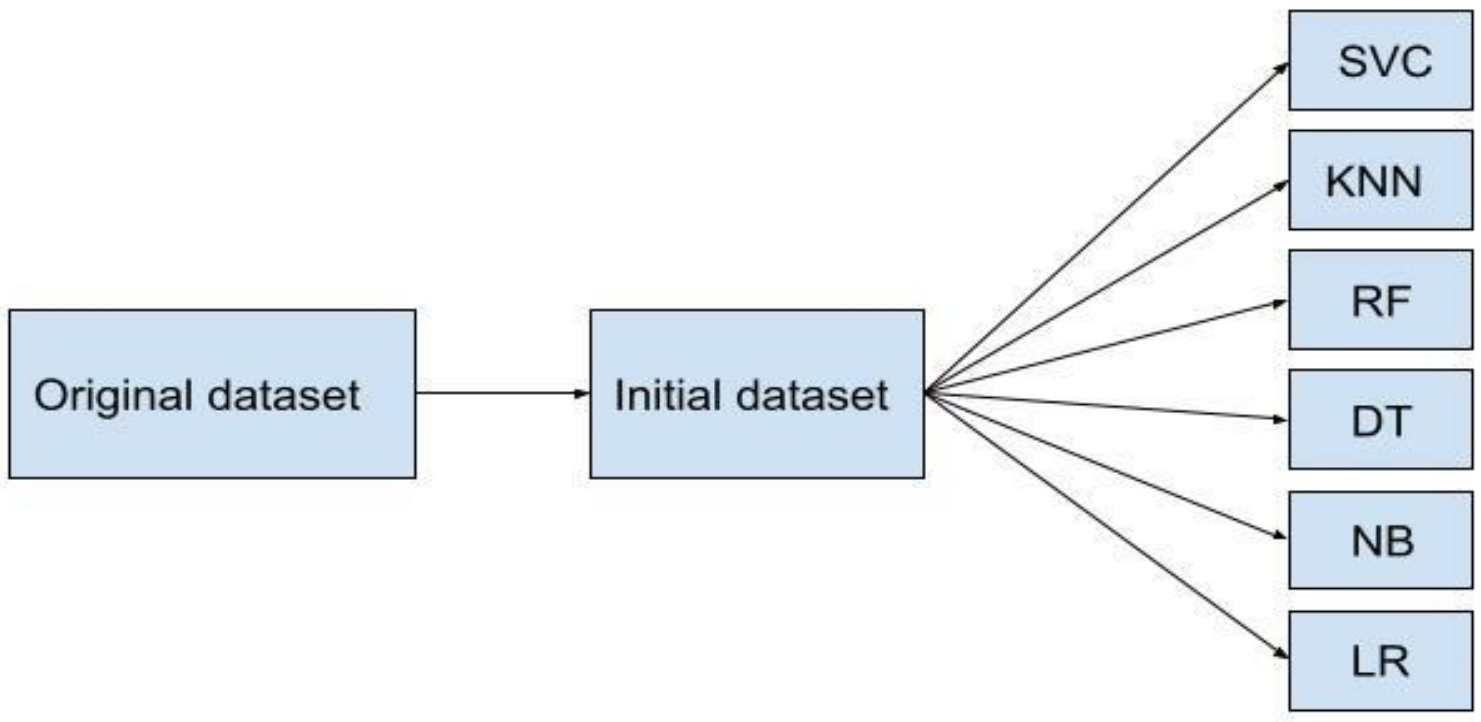

Figure 3: Dataset cleaning framework

## A. First step–Data processing using general techniques of NLP

Each dataset requires preprocessing, cleaning, and transforming to make the dataset appropriate for machine learning models. See Figure 3 for the data cleaning approach using six machine learning algorithms. We have used Python's NLTK library to clean both datasets. Then, we delete irrelevant numbers, stop words, and punctuation from the dataset, lower the case, and lemmatise these data. We named this dataset as an initial dataset. Finally, we used TF-IDF vectorisation to transform both datasets' input classes. After the vectorisation, we feed the data into the machine learning model.

## B. The second step is to create a baseline model and record the baseline performance.

After completing the dataset cleaning, using the general processes of NLP, on both datasets, both datasets were split into train and test sets using Sklearn's train_test_split. Our training set comprises 80% of the dataset, while the test set comprises 20%. The class was again imbalanced during splitting, so we used Sklearn's SMOTE library to synthesise and balance the dataset. Sklearn was used to import the six models, and each model was trained, and the performance was recorded for each model. We recorded each model's accuracy, F-1 score, recall score, precision score, roc_auc score, confusion matrix, and Roc_Auc curve. This was our baseline performance on the original dataset.

## C. The third step is cleaning the original dataset using the ensemble technique.

After observing the performance of each model, which was relatively low for each model, the low performance occurred due to anomalies in the original dataset. This anomaly occurs due to incorrect classification or outliers in the original dataset. Therefore, an ensemble technique was used to remove those anomalies in the original dataset, which was propagated to our initial dataset. In the ensemble (see Figure 4), we used previously trained models for voting purposes. In this case, all previously trained models act as domain experts.

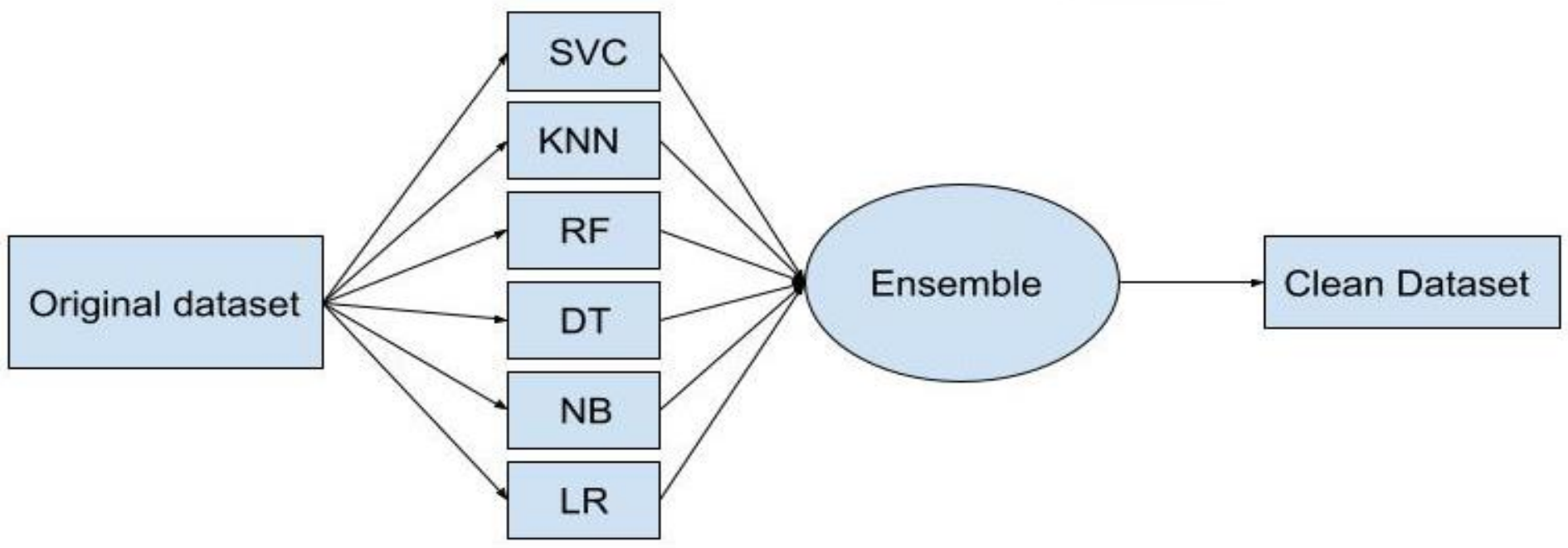

Figure 4: Ensemble Technique is described below.

From the original dataset, 4-class data points are taken and cleaned using the traditional NLP technique. The whole 4-class dataset was predicted individually with those previously trained six models. Moreover, the individual prediction is stored in six individual prediction arrays.

A bias array is created based on the experience of their previous performance (baseline performance). That is, the best-performing model gets a better bias value. For example, we have set bias = [1.0, 1.3, 0.7, 0.9, 1.1, 1.2] for our six algorithms. The biasing process was valid when the vote was three by three on a specific classification.

We have initiated ensenble_prediction = []. Where the combination (voting) of those six algorithm output classes will be stored for the whole dataset. We loop through the prediction array for each model. Within the loop, for each prediction, we have initiated an array predicted_class = [0, 0, 0, 0] and updated the predicted_class index value with those six algorithms' predictions, plus adding their bias values. We take the max (predicted_class) for each prediction. Thus, we find the maximum voting result. Moreover, it is assigned to the ensemble prediction.

Finally, we exclude the data points from the original dataset whose class does not match our ensemble prediction array. Thus, we find the improved dataset.

### D. The fourth step is to find the performance of the improved dataset

In this step, we took our improved dataset, followed the first two steps, and recorded the accuracy, F-1 score, recall score, precision score, Roc_Auc score, confusion matrix, and Roc_auc curve on the improved dataset. These were our performance measurements after the ensemble occurred in the original dataset.

## 4. Performance metrics for the experiments

This research uses several performance measures for machine learning classification models to assess how well the algorithms perform in this experiment. The following performance metrics are considered, including accuracy (AC), precision, recall, F1-score, confusion matrix (CM), and the receiver operating characteristic (ROC) plot. Those metrics also involve the variables of true positive (TP), true negative (TN), false positive (FP), and false negative (FN).

Accuracy is one metric for evaluating classification models. The accuracy is the fraction of the predictions our model got right.

Accuracy is the ratio of correctly labelled images to the total number of samples (Kawasaki et al., 2015, December).

$$Accuracy = \frac{TP + TN}{TP + FP + TN + FN}$$

Precision is defined as the probability given a positive label and how many are positive [56]. Precision tells us how many of the correctly predicted cases turned out to be positive. Precision is a valuable metric in cases where FP is a higher concern than FN.

$$Precision = \frac{TP}{TP + FP}$$

Recall or sensitivity is the accuracy of positively predicted instances describing how many were labelled correctly [55]. Recall tells us how many positive cases we could predict correctly with our model. The Recall is a valuable metric in cases where FN trumps FP.

$$F1 - score = \frac{2}{\frac{1}{Recall} + \frac{1}{Precision}}$$

The F1 score, as an additional measure of classification accuracy, considers both precision and recall. F1-score is a harmonic mean of precision and recall, giving a combined idea about these two metrics. It is maximum when the precision is equal to recall.

F1= 2*(precision* recall)/ Precision + recall

Specificity indicates what percentage of those who do not have the condition have an adverse result on the test. A particular test is good at excluding most people who do not

$$Specificity = \frac{TN}{TN + FP}$$

have the condition. Therefore, a positive test result can confidently rule out the condition for a particular individual. The equation is given below:

A confusion matrix (CM) is a table format that allows the performance evaluation of an algorithm. CM visualises important predictive analytics like recall, specificity, accuracy, and precision. Confusion matrices are helpful because they directly compare values like TP, FP, TN, and FN.

Lastly, the ROC curve measures the discriminatory capacity of classification models. The ROC curve is an evaluation metric for binary classification problems. It is a probability curve that plots the TPR against FPR at various threshold values and essentially separates the 'signal' from the 'noise'. The Area Under the Curve (AUC) measures a classifier's ability to distinguish between classes and is used as a summary of the ROC curve. The higher the AUC, the better the model's performance in distinguishing between the positive and negative classes (Jiménez-Valverde, 2012).

# 5. Results of the experiments

The results of the experiments are described below:

## E. Results of the experiment – 1

The resultant impacts of the ensemble model in this research are removing wrong data points, improved dataset quality, and improved accuracy. The result analysis of the ensemble model starts with the wrong data point removing:

Six models (Naive Bayes, KNN, SVC, Decision Tree, Random Forest, and Logistic Regression) were used to understand the baseline performance on both datasets. These performances were recorded and observed in various machine learning metrics such as accuracy, F-1 score, recall score, precision score, Roc_Auc score, confusion matrix, and roc_auc curve. These baseline performances were lower for each model in both datasets as the original dataset contains anomalies such as incorrect classification in some records, which were propagated to the initial dataset regardless of classical NLP data cleaning before training. Incorrect classification reduces the performance of each model. Therefore, anomalies must be removed before generating the initial dataset through classical NLP data cleaning. However, we need a domain expert (Human) who can manually check each record and correct or remove the incorrect classification. Unfortunately, the task is quite cumbersome for the domain expert and unfeasible in the modern era of computing. Therefore, as the technology progresses and data size increases daily, it demands increasingly automated data cleaners, in this case, domain experts. In this research, we used six models as domain experts to automate the cleaning process. Six models were used to make the ensemble(voting) for determining and removing the anomalies from the original dataset more reliable and robust.

A dataset containing a wrong data point or anomaly will perform poorly on machine learning models. In our experiment, the ensemble-based cleaning methods could detect and remove the anomalies from the original dataset because the ensemble(voting) with six models could detect the wrong data points. For example, suppose a record was assigned a class 0 in the original dataset, and the ensemble(voting) suggests the record belongs to class 1 (that means most models vote for class 1 for that particular record). In that case, it suggests the record contains incorrect data points. Therefore, we removed all such records from the original dataset. Finally, when we import new models to understand the newly created datasets' performance, we find each model's improved performance on each machine learning performance metric. Although ensemble cleaning reduces the original dataset size by removing the anomalies, we found better performance on the shorter datasets, which asserts the statement about the dataset's quality over the dataset's size. An experiment on the two datasets shows that the framework is reliable and thus can be applied to various datasets.

Our experimental results using a single ML (Naive et al.) suggested that low accuracy was due to the presence of wrong classification or wrong data points. An example of a wrong classification/wrong data point is given below:

| **Class** | **Description** |
|---|---|
| Cardiovascular | *"PREOPERATIVE DIAGNOSIS: ,Wrist ganglion.<br><br>POSTOPERATIVE DIAGNOSIS: Wrist ganglion.<br><br>TITLE OF PROCEDURE: Excision of the dorsal wrist ganglion.<br><br>PROCEDURE: After administering appropriate antibiotics and general anaesthesia, the upper extremity was prepped and draped in the usual standard fashion. The arm was exsanguinated with an Esmarch, and a tourniquet inflated to 250 mmHg* was." |

Table 1: Wrong datapoint

The wrong data point (see Table 1) is a data point that belongs to one class but is labelled by mistake as belonging to the other class. The descriptions suggest that it has no relation to the cardiovascular/pulmonary. Therefore, to eliminate such data, we applied the ensemble technique. However, we trained the dataset using six algorithms using the ensemble technique. As predicted by the above description, the training model should be used in surgery. In Table 2, the data points are correctly classified after the ensemble module. However, the ensemble model was able to classify the wrong data point, which allowed to move the data to the correct data point; an example is given below:

| Class | Description |
|---|---|
| Cardiovascular | *"Adenosine with the nuclear scan as the patient cannot walkon a treadmill.<br><br>INTERPRETATION: A resting heart rate of67 and blood pressure of 129/86. EKG, normal sinus rhythm. Post-Lexiscan 0.4 mg, heart rate was 83, blood pressure 142/74. EKG remained the same. No symptoms were noted.<br><br>SUMMARY: 1. Nondiagnostic adenosine stress test.2. Nuclear interpretation as below.<br><br>NUCLEAR INTERPRETATION: Resting and stress images were obtained with 10.4, 33.1 mCi of"* |

Table 2: Correct data point

However, the results of the experiments are described below:

Results are in percentage (%)
BE- Result before ensemble AE- Result ensemble

| | **Naive Bayes** | | **KNN** | | **SVC** | | **Decision Tree** | | **Random Forest** | | **Logistic Regression** | |
|---|---|---|---|---|---|---|---|---|---|---|---|---|
| *Metrics* | BE | AE | BE | AE | BE | AE | BE | AE | BE | AE | BE | AE |
| Accuracy | 75 | 88 | 74 | 83 | 73 | 94 | 56 | 78 | 68 | 88 | 73 | 93 |
| F1-Score | 74 | 89 | 73 | 83 | 73 | 94 | 56 | 78 | 68 | 88 | 73 | 93 |
| Recall | 75 | 88 | 74 | 82 | 74 | 94 | 56 | 78 | 68 | 88 | 73 | 93 |
| Precision | 78 | 91 | 75 | 89 | 73 | 94 | 57 | 79 | 68 | 88 | 73 | 93 |
| ROC Auc | 93 | 98 | 90 | 95 | 90 | 93 | 71 | 85 | 88 | 98 | 93 | 99 |

Table 3: Results obtained from the experiments

The results of accuracy, F1-score, recall, precision, and ROC accuracy of the classes (see Table 3) suggest that after the voting-based ensemble technique, the dataset was cleaner than the original dataset. The accuracy, F1-score, recall, and precision improvement was around 10% to 20%. Moreover, the highest ROC accuracy of the classes was 14% (decision tree algorithm).

The confusion matrix displays the number of predicted data

OD= Orthopedic SR= Surgery

CV= Cardiovascular CT= Consultation

**Before ensemble voting** **After ensemble voting** **Before ensemble voting** **After ensemble voting**

Naive Bayes

| | Before SR | Before OR | Before CT | Before CV | After SR | After OR | After CT | After CV |
|---|---|---|---|---|---|---|---|---|
| CV | 35 | 25 | 0 | 5 | 43 | 12 | 0 | 6 |
| CT | 1 | 69 | 2 | 0 | 9 | 77 | 1 | 0 |
| OR | 0 | 10 | 63 | 2 | 0 | 6 | 54 | 0 |
| SR | 6 | 0 | 20 | 48 | 2 | 0 | 1 | 51 |

KNN

| | Before SR | Before OR | Before CT | Before CV | After SR | After OR | After CT | After CV |
|---|---|---|---|---|---|---|---|---|
| CV | 40 | 20 | 0 | 3 | 52 | 8 | 0 | 1 |
| CT | 2 | 67 | 3 | 0 | 0 | 77 | 1 | 0 |
| OR | 0 | 6 | 56 | 13 | 0 | 11 | 49 | 0 |
| SR | 8 | 2 | 17 | 47 | 1 | 18 | 2 | 33 |

SVC

| | Before SR | Before OR | Before CT | Before CV | After SR | After OR | After CT | After CV |
|---|---|---|---|---|---|---|---|---|
| CV | 51 | 7 | 0 | 5 | 53 | 3 | 0 | 5 |
| CT | 7 | 59 | 6 | 0 | 1 | 76 | 1 | 0 |
| OR | 0 | 6 | 54 | 15 | 0 | 3 | 57 | 0 |
| SR | 10 | 0 | 20 | 44 | 1 | 0 | 1 | 52 |

Decision Tree

| | Before SR | Before OR | Before CT | Before CV | After SR | After OR | After CT | After CV |
|---|---|---|---|---|---|---|---|---|
| CV | 34 | 13 | 9 | 7 | 47 | 5 | 1 | 8 |
| CT | 13 | 49 | 10 | 0 | 10 | 64 | 3 | 1 |
| OR | 2 | 11 | 39 | 23 | 3 | 4 | 42 | 11 |
| SR | 11 | 0 | 25 | 38 | 7 | 0 | 2 | 45 |

Random Forest

| | Before SR | Before OR | Before CT | Before CV | After SR | After OR | After CT | After CV |
|---|---|---|---|---|---|---|---|---|
| CV | 39 | 14 | 0 | 10 | 46 | 6 | 0 | 9 |
| CT | 5 | 62 | 5 | 0 | 2 | 74 | 2 | 0 |
| OR | 0 | 7 | 49 | 19 | 0 | 5 | 51 | 4 |
| SR | 10 | 0 | 19 | 45 | 2 | 0 | 2 | 50 |

Logistic Regression

| | Before SR | Before OR | Before CT | Before CV | After SR | After OR | After CT | After CV |
|---|---|---|---|---|---|---|---|---|
| CV | 50 | 7 | 0 | 6 | 53 | 3 | 0 | 5 |
| CT | 7 | 59 | 6 | 0 | 2 | 75 | 1 | 0 |
| OR | 0 | 5 | 54 | 16 | 0 | 4 | 56 | 0 |
| SR | 10 | 0 | 20 | 44 | 2 | 0 | 1 | 51 |

Figure 5: Experiment 1- Confusion matrix of before ensemble and after ensemble

The confusion matrices of six algorithms before and after the ensemble voting were obtained to get a clearer picture. Six algorithms are used. Consequently, twelve (12) matrixes are produced. Figure 5 depicts that the four classes, Orthopaedic (OD), Surgery (SR), Cardiovascular (CV), and Consultation (CT) prediction capabilities of the algorithms used in this study have been increased. This is because the voting system used by the ensemble has positioned the data points to the correct class. Not only improving the number of true positives (TP), but also the number of true negatives (TN), false positives (FP), and false negatives (FN) were reduced. In general, logistic regression has produced a better result, which, in this case, is a true positive. However, overall, the performance of the surgery class improved by all applied algorithms.

Naive Bayes

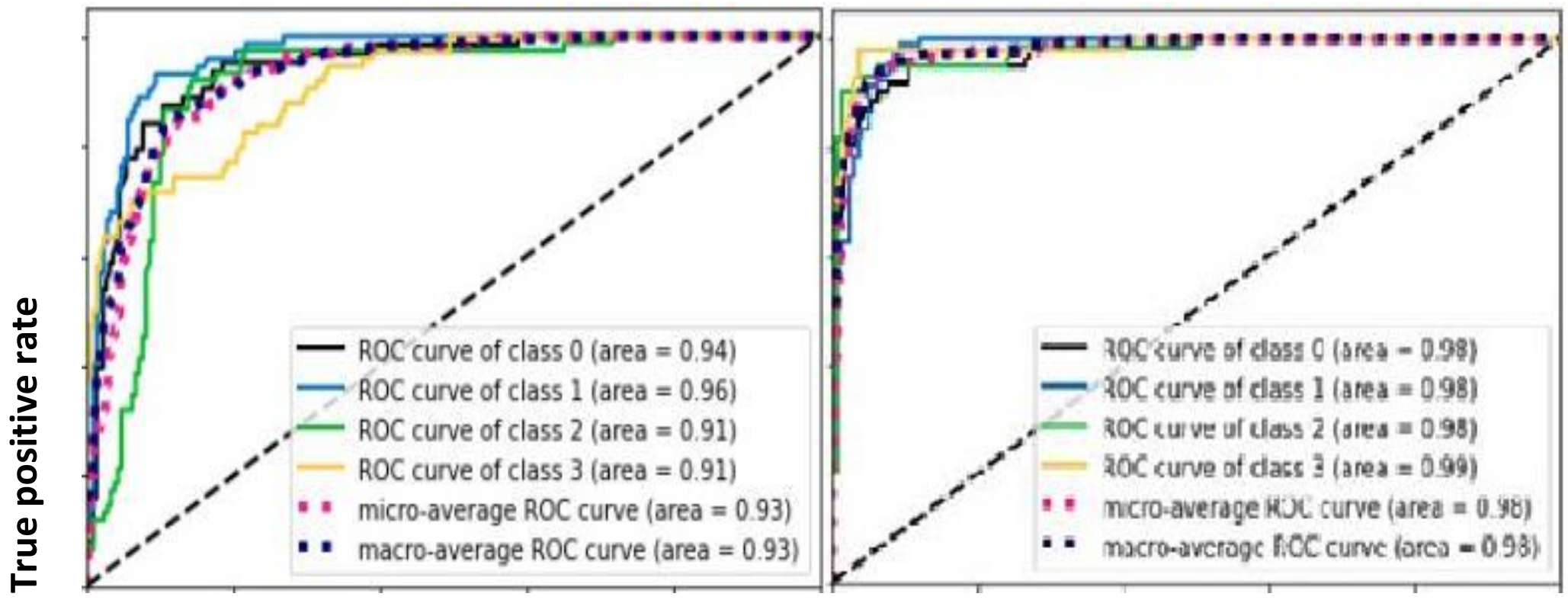

KNN

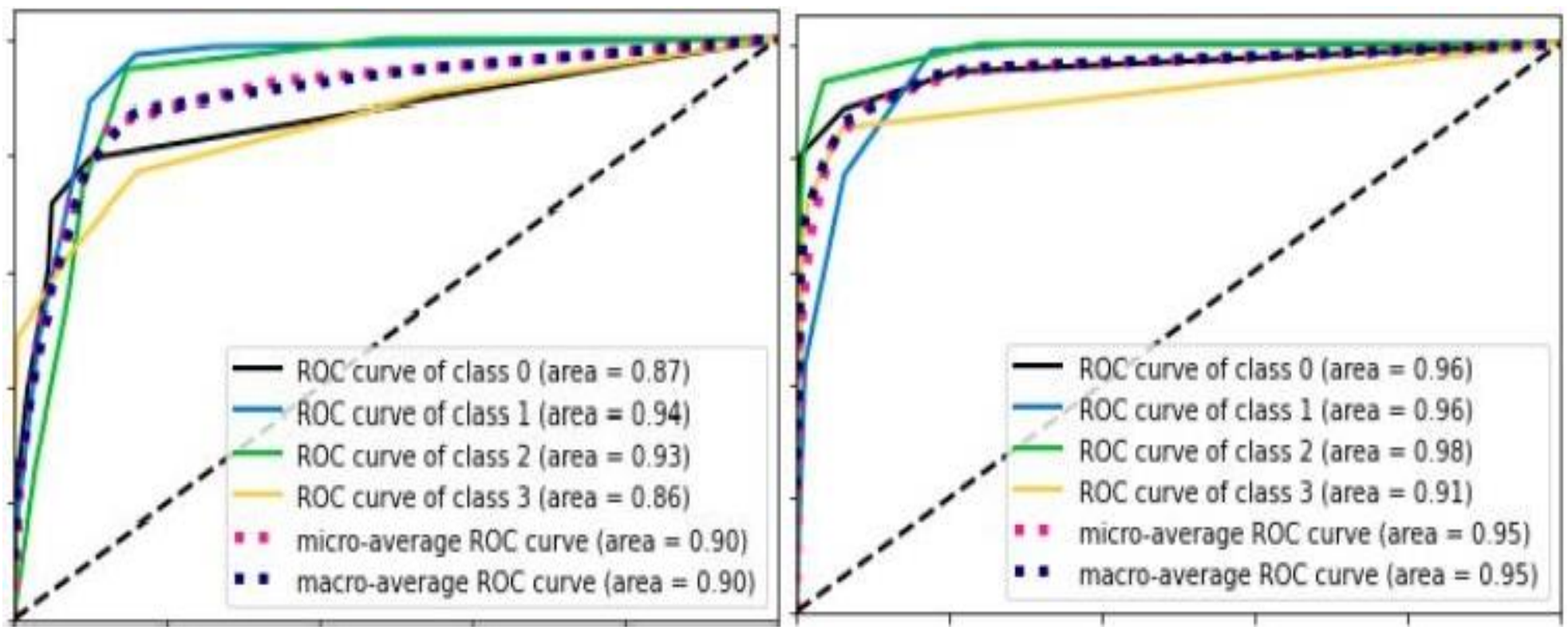

SVC

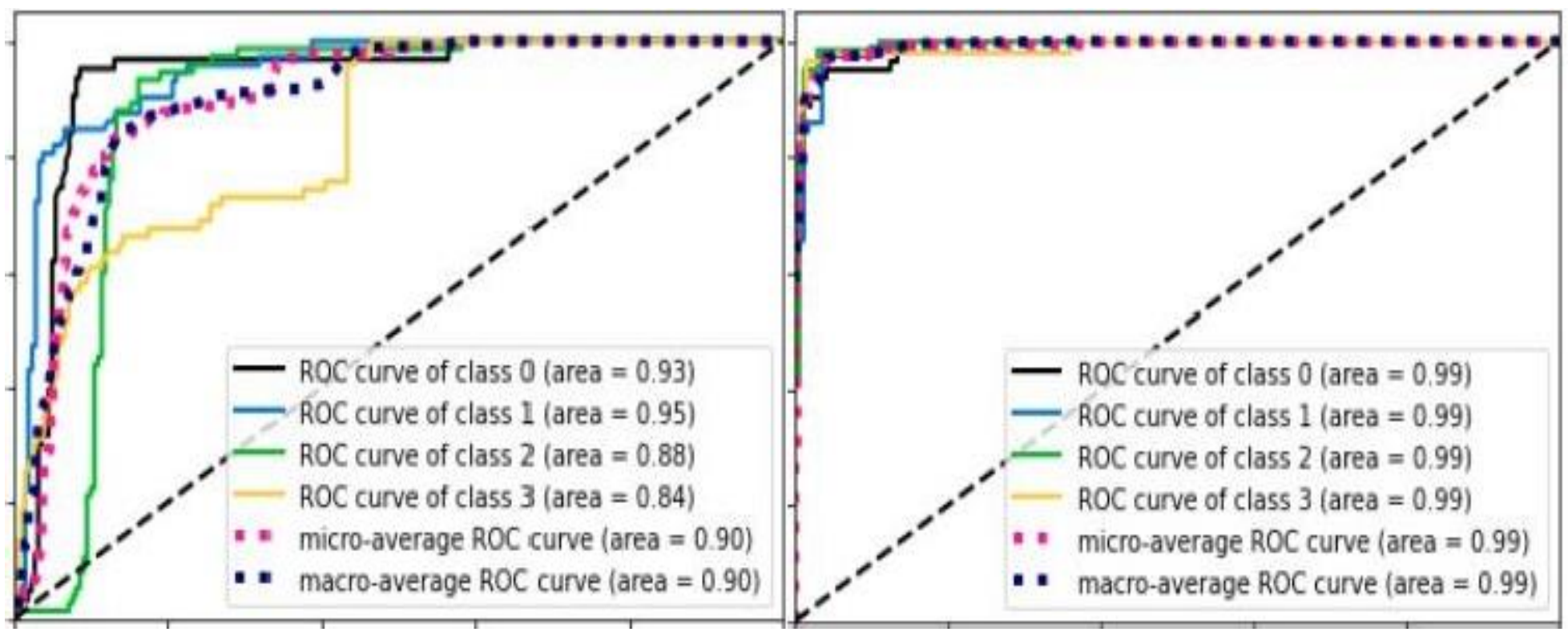

Decision Tree

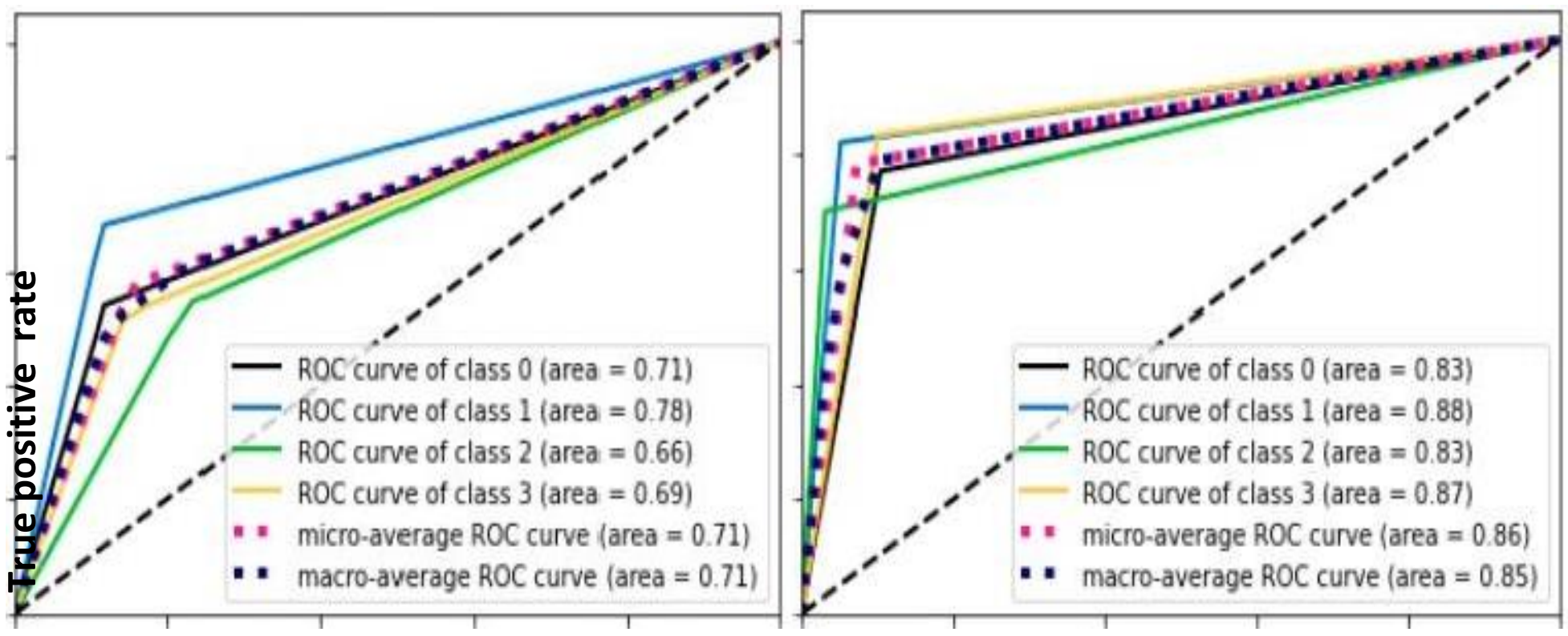

Random Forest

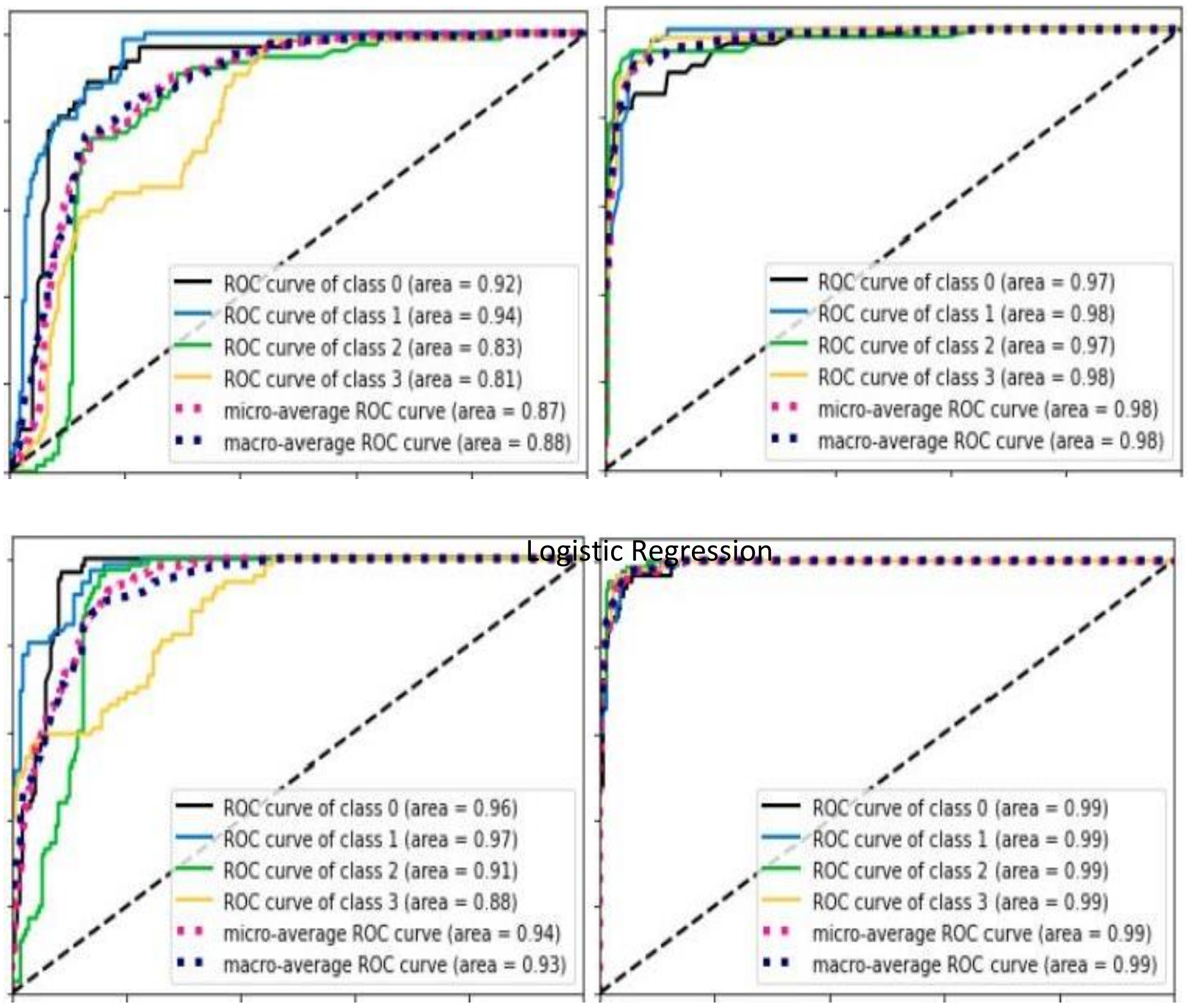

**False positive rate**

Figure 6: ROC of algorithms

The ROC curve is plotted (Figure 6) for a graphical understanding of the performance of the algorithms used for the diagnostic ability of classes. In the ROC curve, the y-axis represents the sensitivity or TPR, and the X-axis represents specificity or FPR. A perfect ROC curve is expected to move along up, following the line of sensitivity and then across to the right towards the specificity. The closer the curve comes to the 45-degree diagonal of the ROC.

Space, the less accurate the test. Following the application of ROC, is it visible that the six algorithms were able to improve the accuracy by applying the ensemble-based voting for cleaning?

## F. Results of the experiment – 2

As stated earlier, the second experiment included a dataset that contains 53933 data points and four classes: prescription (14500), appointments (12960), ask_a_doctor (11743), and miscellaneous (10460) used in the experiments and as each class is balanced to 10460 data points. However, the results of the experiments are described below:

Results are in percentage (%)
BE- Result before the ensemble
AE- Result after ensemble

| | **Naive Bayes** | | **KNN** | | **SVC** | | **Decision Tree** | | **Random Forest** | | **Logistic Regression** | |
|---|---|---|---|---|---|---|---|---|---|---|---|---|
| *Metrics* | BE | AE | BE | AE | BE | AE | BE | AE | BE | AE | BE | AE |
| Accuracy | 74 | 76 | 64 | 65 | 76 | 78 | 70 | 72 | 75 | 77 | 75 | 77 |
| F1-Score | 74 | 75 | 63 | 65 | 76 | 78 | 70 | 72 | 75 | 76 | 75 | 77 |
| Recall | 74 | 76 | 64 | 65 | 76 | 78 | 70 | 72 | 75 | 76 | 75 | 77 |
| Precision | 74 | 76 | 64 | 66 | 76 | 78 | 70 | 72 | 75 | 76 | 75 | 77 |
| ROC Auc | 91 | 92 | 83 | 84 | 92 | 93 | 82 | 83 | 91 | 92 | 91 | 93 |

Table 4: Results obtained from the experiments

The second experiment's results of accuracy, F1-score, recall, precision, and ROC accuracy of the classes Prescription, Appointment, Ask_a_Doctor, and Miscellaneous (see Table 4) support experiment 1 that after the voting-based ensemble technique, the dataset was cleaner than the original dataset. However, the accuracy, F1-score, recall, and precision improvement are less in a larger dataset.

The confusion matrix displays the number of predicted data

*PS= Prescription* *AP=Appointment* *AS= Ask_a_Doctor* *MS= Miscellaneous*

SVC

| | Before ensemble voting | | | | After ensemble voting | | | |
|---|---|---|---|---|---|---|---|---|
| PS | 1608 | 195 | 188 | 25 | 1749 | 156 | 168 | 11 |
| AP | 129 | 1454 | 343 | 209 | 124 | 1377 | 285 | 220 |

Decision Tree

| | Before ensemble voting | | | | After ensemble voting | | | |
|---|---|---|---|---|---|---|---|---|
| PS | 1572 | 209 | 196 | 39 | 1671 | 199 | 163 | 51 |
| AP | 232 | 1272 | 396 | 235 | 184 | 1229 | 329 | 264 |

Naive Bayes

| | Before ensemble voting | | | | After ensemble voting | | | |
|---|---|---|---|---|---|---|---|---|
| PS | 1608 | 167 | 204 | 37 | 1739 | 147 | 176 | 22 |
| AP | 171 | 1364 | 324 | 276 | 177 | 1254 | 302 | 273 |
| AS | 175 | 205 | 1555 | 164 | 151 | 175 | 1351 | 94 |
| MS | 66 | 180 | 184 | 1688 | 61 | 168 | 177 | 1695 |

KNN

| | Before ensemble voting | | | | After ensemble voting | | | |
|---|---|---|---|---|---|---|---|---|
| PS | 1564 | 168 | 246 | 38 | 1561 | 207 | 293 | 23 |
| AP | 464 | 974 | 416 | 281 | 161 | 1094 | 522 | 229 |
| AS | 323 | 236 | 1372 | 168 | 111 | 430 | 1147 | 83 |
| MS | 176 | 207 | 293 | 1442 | 58 | 312 | 403 | 1328 |

| | | | | | | | | | | | | | | | | |
|---|---|---|---|---|---|---|---|---|---|---|---|---|---|---|---|---|
| AS | 97 | 253 | 1636 | 113 | 112 | 201 | 1376 | 82 | 179 | 311 | 1465 | 144 | 168 | 260 | 1226 | 117 |
| MS | 40 | 203 | 191 | 1684 | 43 | 204 | 154 | 1700 | 63 | 260 | 208 | 1587 | 54 | 262 | 177 | 1608 |

Random Forest

Logistic Regression

| | | | | | | | | | | | | | | | | |
|---|---|---|---|---|---|---|---|---|---|---|---|---|---|---|---|---|
| PS | 1618 | 195 | 175 | 28 | 1744 | 152 | 159 | 29 | 1608 | 181 | 191 | 36 | 1729 | 146 | 184 | 25 |
| AP | 169 | 1410 | 324 | 232 | 130 | 1330 | 295 | 251 | 151 | 1406 | 344 | 234 | 146 | 1321 | 314 | 225 |
| AS | 124 | 290 | 1558 | 127 | 126 | 235 | 1309 | 101 | 132 | 222 | 1603 | 142 | 122 | 177 | 1379 | 93 |
| MS | 43 | 196 | 174 | 1705 | 46 | 199 | 168 | 1688 | 52 | 203 | 181 | 1682 | 40 | 197 | 169 | 1695 |
| | MS | AS | AP | PS | MS | AS | AP | PS | MS | AS | AP | PS | MS | AS | AP | PS |

Figure 7: Experiment 2- Confusion matrix of before ensemble and after ensemble

The confusion matrix (see Figure 7) of six algorithms before and after the ensemble voting system improves the number of true positives (TP), and the true negative (TN), false positive (FP), and false negative (FN) were reduced. In general, logistic regression has produced a better result, which, in this case, is a true positive. However, overall, the performance of the surgery class improved by all applied algorithms.

*Class 0= Prescription* *Class 1=Appointment* *Class 2= Ask_a_Doctor* *Class 3= Miscellaneous*

Naive Bayes

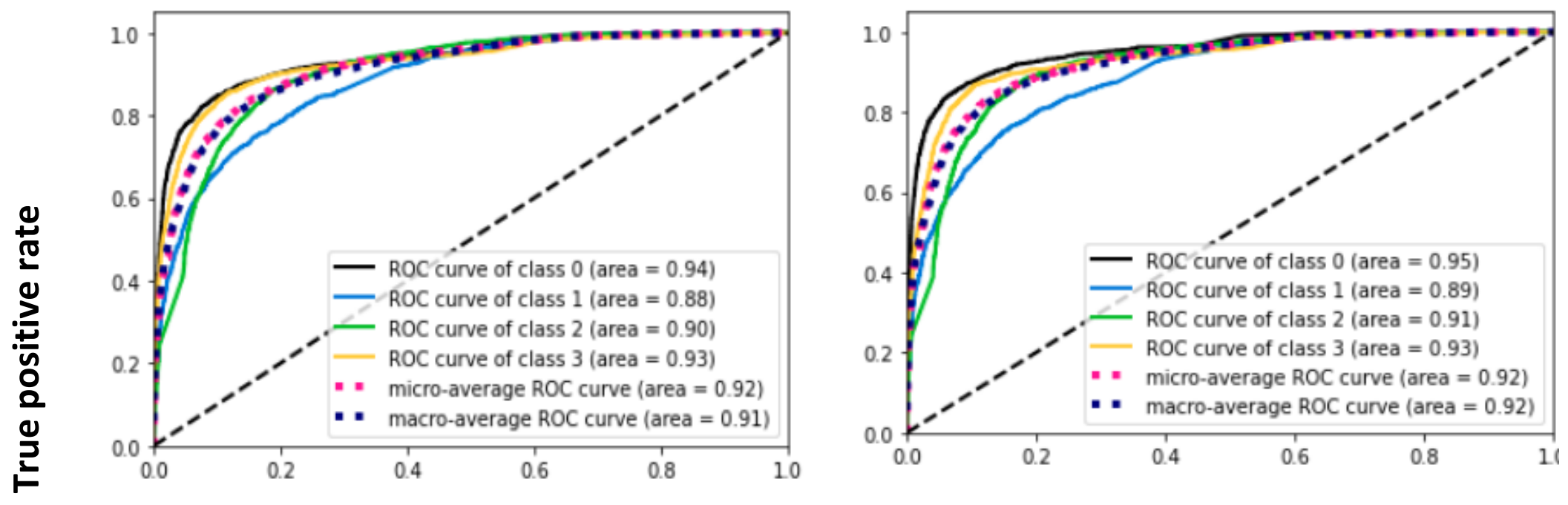

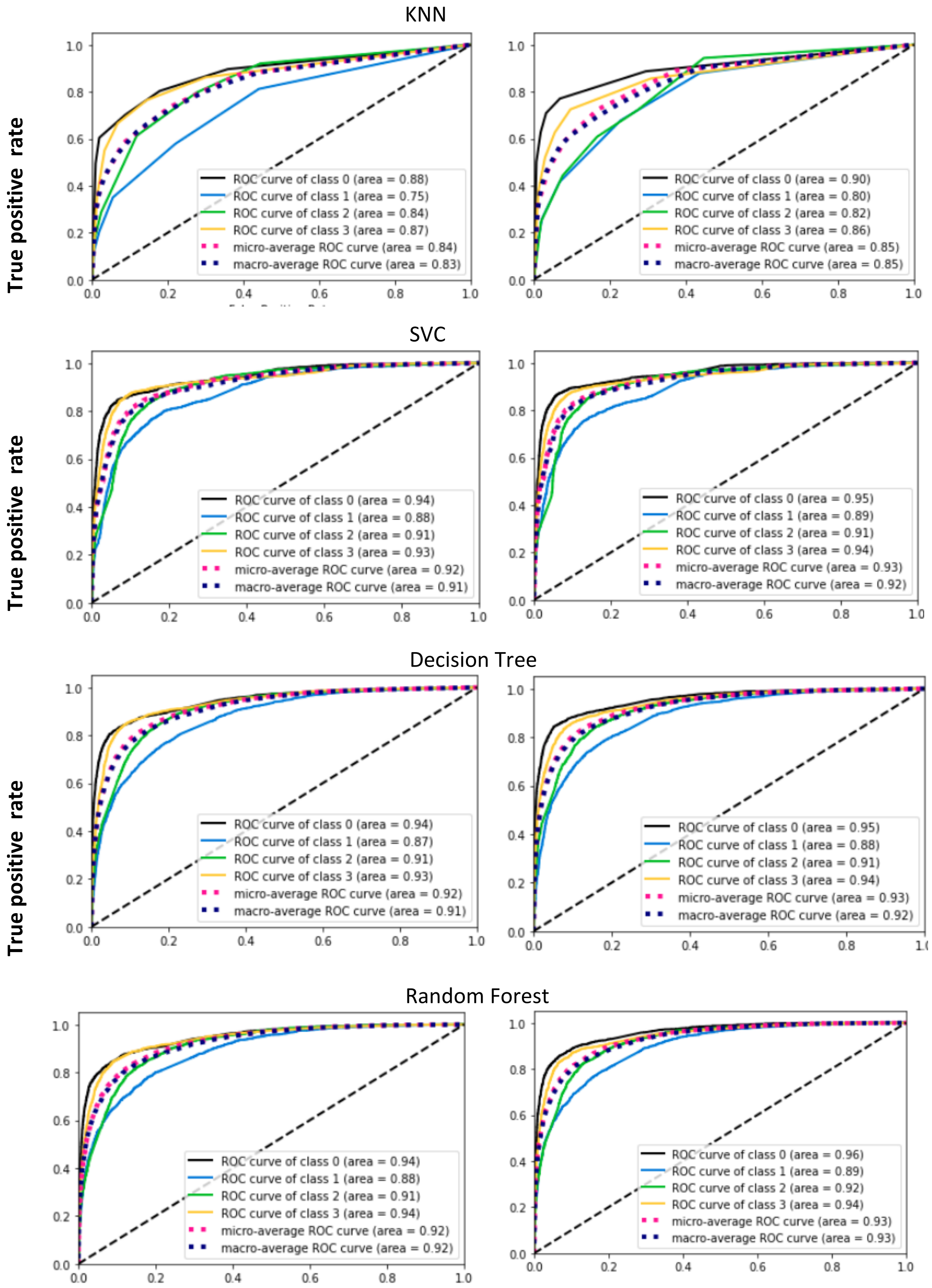

Logistic Regression

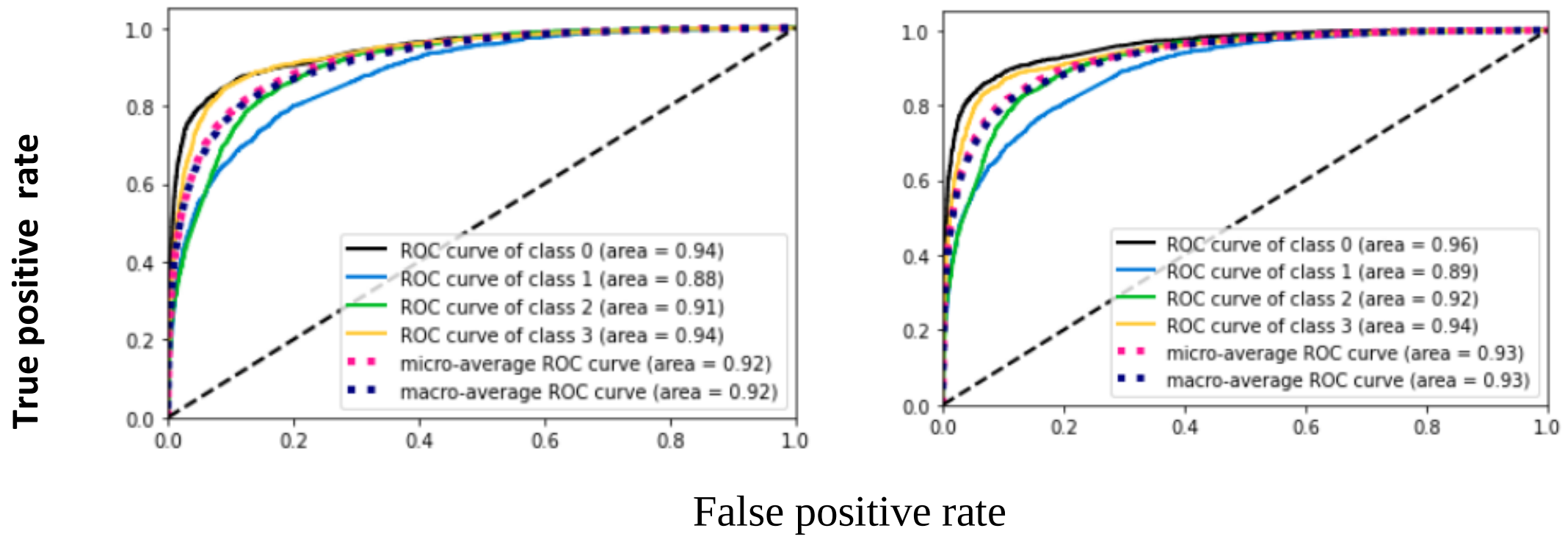

Figure 8: ROC of algorithms

The ROC curve is plotted (Figure 8) for a graphical understanding of the performance of the algorithms used for the diagnostic ability of classes. In the ROC curve, the y-axis represents the sensitivity or TPR, and the X-axis represents specificity or FPR. A perfect ROC curve is expected to move along up, following the line of sensitivity and then across to the right towards the specificity. The closer the curve comes to the 45-degree diagonal of the ROC space, the less accurate the test is. Following the application of ROC, it is visible that the six algorithms improved the accuracy by applying ensemble-based voting for cleaning in small and large datasets.

# 6. Contribution

The study offers several critical contributions. Some contributions are applied that can benefit data scientists, especially NLP programmers, medical practitioners, and the ICT industry, while some are from academia and of significance to the research community.

Researchers are increasingly interested in investigating the opportunities to improve the data cleaning process in NLP to increase prediction and provide inaccurate business logic (Yang et al., 2021b). However, data variety is the main obstacle in the patient management system (Deshpande et al., 2020). Moreover, the data-cleaning process is complex and time-consuming. Therefore, as the amount of data is increasing daily, cleaning data quickly and efficiently is becoming critical. Research suggests that data scientists spend sixty per cent (60%) of their time during data cleaning when processing large amounts of data CrowdFlower (2016). Data cleaning in NLP is an interesting problem-solving tool for data in this significant data era.

The importance of data cleaning has been addressed by several scholars, for example, (Bertsimas & Wiberg 2020; Suresh et al., 2020; Tran et al., 2017; Mohamed et al., 2011; Yang et al., 2021a). The study by Tran et al. (2017) purported that patient data cleaning is an important quality.

The study by Tran et al. (2017) purported that patient data cleaning is a necessary quality assurance of patient health information. Moreover, Mohamed et al. (2011) advocated that the accuracy of patient data provides reliable figures, which are essential for patients, doctors, and hospitals.

A scientific method is the experimental method. It is future-oriented because the researcher is attempting to assess something novel. It is a process of adding to a body of information that has already been accumulated. As a result, the researcher starts with the premise that the research setting he wants to analyse has never existed and will never exist. The term ”situation” refers to both a program, curriculum, or strategy for organising classes, as well as a ”situation” constructed to put students through their paces Nayak & Singh (2021).

The study offers several critical contributions. Some contributions are applied that can benefit the data scientists, especially the NLP programmers, medical practitioners and the ICT industry, while some are from academia and of significance to the research community.

Researchers are increasingly interested in investigating the opportunities to improve the Data cleaning process in NLP to increase prediction and provide inaccurate business logic Yang et al. (2021b). However, data variety is the main obstacle in patient management systems Deshpande et al. (2020). Moreover, the data-cleaning process is complex and time-consuming. Therefore, as the amount of data is increasing daily, it is essential to clean data quickly and efficiently. Research suggests data scientists spend sixty per cent (60%) of their time during data cleaning when processing large amounts of data CrowdFlower (2016). Data cleaning in NLP is an exciting problem for data in this significant data era.

The importance of data cleaning has been addressed by several scholars, for example, Yang et al. (2021a), Bertsimas & Wiberg (2020), Suresh et al. (2020), Tran et al. (2017) and Mohamed et al. (2011). The study by Tran et al. (2017) purported that patient data cleaning is a necessary quality assurance of patient health information. Moreover, Mohamed et al. (2011) advocated that the accuracy of patient data provides reliable figures; these figures are significant to patients, doctors and hospitals.

This study focuses on improving the patient dataset so that valuable information can be retrieved. From this perspective, this research develops mid-range theory- the theory is that the ensemble technique in NLP has an opportunity to improve patient-related information. Gregor (2006)

[56] defines midrange theory as a moderately abstract theory with a narrow scope that may readily lead to testable hypotheses. Midrange theory is essential because it deals with practice-based disciplines, notably in the social sciences (Haxeltine et al.,2017), and it is also highlighted that social phenomena are made up of social interactions and practices that are profoundly anchored in technology (Haxeltine et al.,2017). As an area of technology, NLP is also related to organisational business process management and social activities in a specific setting.

Researchers should also focus on methodological advancement, according to Barki (2008). He claimed that the study had insufficient guidelines for construct creation. Following these ideas, this thesis provides a well-documented research journey outlining the choice of the NLP basic model. The research methodology chapter (chapter 5) describes the research model development activities, justification of the qualitative research methodology and an experimental approach. Finally, this study needs a thorough literature review of NLP to understand the current knowledge gap.

The study's ensemble technique in the NLP data cleaning method is projected to have significant practical significance in machine learning, data science, the NLP domain, and the Bangladesh government. The Bangladesh government has set 2030 as a deadline for achieving Sustainable Development Targets (SDGs), and this study's ensemble technique in NLP data cleaning method SDG goals. Bangladesh is likely one of the countries that need the most research in the machine learning, data science, and NLP domains. Because the current Bangladeshi government is making many attempts to improve healthcare through the digitisation of the hospitals, this study may give insight into how patient records can be improved through ensemble techniques in the NLP data cleaning method.

# 7. Future Work and Conclusion

Like many other research studies, this research also has several limitations. Firstly, the data used in the experiment was secondary. The investigators approached several hospitals and clinics; however, it was impossible to apply primary data because of data security. Although the accuracy rate was improved by applying the ensemble technique, there might be a more efficient method than we presented in this research. Therefore, our accuracy in the given dataset is limited. We do not claim that the accuracy will also persist in other datasets. One can criticise using a small dataset whose findings may not be generalisable. I recognise that this study used a small dataset, but a more extensive dataset will be tested in future. Lastly, this study is liable for the typical limitations of experimental-based research. Experimental research, in general, is conducted in a controlled environment. However, we applied cutting-edge Python packages for data analysis.

However, to our knowledge, this study sets the first foundations for improving clinical datasets using ensemble techniques; patients' medical records are uniquely positioned to uncover the potential for Machine Learning to help patients and healthcare systems. Therefore, the research findings of this study will not only address the call by government and development agencies for the digitisation of Bangladesh and enhance our understanding of NLP capabilities. These findings collectively illustrate NLP's potential role, particularly in patient record management. Furthermore, the study serves as a platform for further investigation to explore the adoption of NLP in the medical sector.

However, to our knowledge, this study sets the first foundations for improving clinical datasets in Bangladesh results using the Ensemble technique. Patient’s medical records are uniquely positioned to uncover the potential for Machine Learning to help patients and healthcare systems. Therefore, the research findings of this study will not only address the call by government and development agencies for the digitisation of Bangladeshi RMG but also enhance our understanding of NLP capabilities. These findings collectively illustrate NLP's potential role, particularly in patient record management. Furthermore, the study serves as a platform for further investigation to explore the adoption of NLP in the medical sector. This contribution is significant as a step forward in Bangladesh's digitisation.

550

555

580

**This study focuses on the improvement of the patient dataset so that valuable information can be retrieved. From this perspective, this research develops mid-range theory- the theory is an ensemble technique in NLP that has an opportunity to improve patient-related information. Gregor (2006).**

### Data Availability Statement

This research is based on the Kaggle dataset. Therefore, no data is available.

### Funding Statement

No funding was received from this received. None of the authors received any funding.